%% file: sample-sigconf.tex
\documentclass[sigconf,screen]{acmart}

\usepackage{graphicx}
\usepackage{amsmath}

\usepackage{amssymb}
\usepackage{dsfont}
\usepackage{wrapfig}
\usepackage{subfig}
\usepackage{multirow}
\usepackage{color, colortbl}
\usepackage{pifont}
\definecolor{LightGray}{gray}{0.9}
\newcommand{\cmark}{\ding{51}}
\newcommand{\xmark}{\ding{55}}
\usepackage{hyperref}

\AtBeginDocument{%
  \providecommand\BibTeX{{%
    \normalfont B\kern-0.5em{\scshape i\kern-0.25em b}\kern-0.8em\TeX}}}

\acmConference[KDD '22]{KDD '22: 28th ACM SIGKDD Conference on Knowledge Discovery and Data Mining}{Aug 14--18, 2022}{Washington DC}
\acmBooktitle{KDD '22: 28th ACM SIGKDD Conference on Knowledge Discovery and Data Mining,
  Aug 14--18, 2022, Washington DC.}
\acmPrice{15.00}
\acmISBN{978-1-4503-XXXX-X/18/06}

\begin{document}

\title{M6-Fashion: High-Fidelity Multi-modal \\ Image Generation and Editing}

\author{Zhikang Li}
\authornote{Both authors contributed equally to this research.}
\author{Huiling Zhou}
\authornotemark[1]
\author{Shuai Bai}
\authornotemark[1]
\affiliation{%
  \institution{Alibaba DAMO Academy}
}
\author{Peike Li}
\authornotemark[1]
\affiliation{%
\institution{Australian Artificial Intelligence Institute, UTS}
\institution{Alibaba DAMO Academy}
}

\author{Chang Zhou}
\author{Hongxia Yang}
\affiliation{%
  \institution{Alibaba DAMO Academy}
}

\renewcommand{\shortauthors}{Alibaba DAMO Academy}

\begin{abstract}
\input{Sections/abstract}
\end{abstract}



\begin{teaserfigure}
\includegraphics[width=\textwidth]{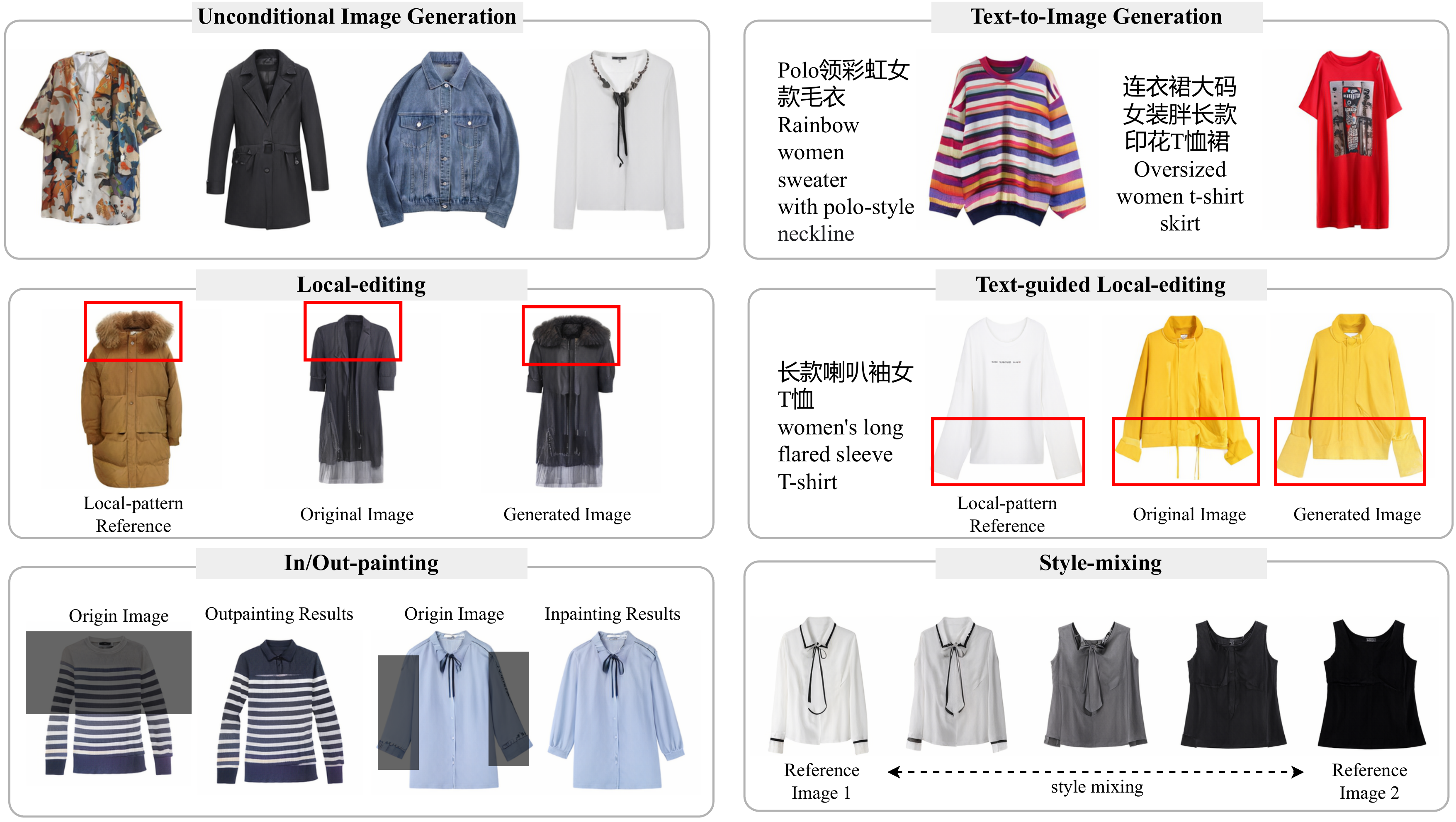}
\caption{The proposed M6-Fashion is a unified, efficient, general framework for multi-modal image generation and editing. M6-Fashion synthesize high-fidelity images for various multi-modal conditional AI-Fashion tasks.
}
\label{fig:teaser}
\vspace{6mm}
\end{teaserfigure}

\maketitle

\section{Introduction}
\input{Sections/1-introduction}

\section{Related Work}
\input{Sections/2-related-work}

\section{Methodology}
\input{Sections/3-method}

\section{Experiment}
\input{Sections/4-experiment}

\section{Conclusion}
\input{Sections/conclusion}


\bibliographystyle{ACM-Reference-Format}
\bibliography{acmart}

\clearpage
\appendix
\input{Sections/appendix}

\end{document}

%% file: Sections/abstract.tex
The fashion industry has diverse applications in multi-modal image generation and editing. 
It aims to create a desired high-fidelity image with the multi-modal conditional signal as guidance.
Most existing methods learn different condition guidance controls by introducing extra models or ignoring the style prior knowledge, which is difficult to handle multiple signal combinations and faces a low-fidelity problem. 
In this paper, we adapt both style prior knowledge and flexibility of multi-modal control into one unified two-stage framework, M6-Fashion, focusing on the practical AI-aided Fashion design.
It decouples style codes in both spatial and semantic dimensions to guarantee high-fidelity image generation in the first stage. 
M6-Fashion utilizes self-correction for the non-autoregressive generation to improve inference speed, enhance holistic consistency, and support various signal controls. 
Extensive experiments on a large-scale clothing dataset M2C-Fashion demonstrates superior performances on various image generation and editing tasks. 
M6-Fashion model serves as a highly potential AI designer for the fashion industry.

%% file: Sections/1-introduction.tex
\input{Tables/method-comparision}





Automatically generating and editing a desired image, given multi-modal control signals, is a challenging yet meaningful task.
Especially in the fashion design area, where there are diverse applications, \textit{e.g.}, style mixing, local editing, text-guided manipulation and text-to-image generation. 
Traditionally, it requires tedious manual operations like sketch drawing and design. With the help of artificial intelligence (AI), it is of great value to make these tasks more human-efficient and user-friendly by providing various manipulation options with simple interaction.
 
Images generated by generative adversarial networks (GAN)~\cite{goodfellow2014generative} have achieved phenomenal fidelity and realism in specific domain. 
As one of the representatives, StyleGAN~\cite{stylegan,karras2021alias} introduces a style-based generator to produce high-resolution images.
Some works~\cite{e4e,psp} learn an encoder to map the intermediate latent space of StyleGAN to achieve meaningful image manipulation. 
However, these GAN-based image synthesis methods~\cite{xia2020tedigan,tao2020df,zhang2017stackgan,li2019controllable} learn different guidance by introducing extra models, which increase the difficulty of training and limit the performance 
under the cross-modal control signals.
Recently, the vector quantization (VQ) based two-stage image synthesis \cite{esser2021taming,razavi2019generating,dalle,ding2021cogview,zhang2021m6} methods have made a great progress. 
The first stage converts images to quantized latent representations with an autoencoder model.
The second stage applies the transformer~\cite{vaswani2017attention} to predict a sequence of tokens, which are converted back to image by the autoencoder in the first stage.
However, these autoencoders typically focus on the reconstruction task in the general domain, ignoring the style prior knowledge of a specific domain. 
It results in the low-fidelity problem for the generated images.
 
In this paper, we take advantage of both style prior knowledge and flexibility of multi-modal control into one unified framework, namely, M6-Fashion.
It models both images and multi-modal control signals uniformly into the sequences of discrete tokens.
Besides simple unconditional generation, it is able to perform various conditional controls, such as style mixing and spatial editing, as shown in Figure~\ref{fig:teaser}.
Together, our proposed two-stage transformer-based M6-Fashion framework servers as a highly potential AI designer for the Fashion industry.
More concretely, in the first stage, we learn the quantized latent representations for converting an image into a sequence of discrete tokens with explicit spatial dimensions by product quantization~\cite{jegou2010product} method. 
Our proposed discrete representation benefits from its spatial-aware characteristic, enabling the generative adversarial networks (GANs) to effectively decouple the style latent code in both spatial and semantic dimension.
In the second stage, a non-autoregressive (NAR) Transformer model is utilized to capture the distribution over sequences of tokens, comprising of tokens from multi-modal control signals and visual tokens from the target image. 
Our NAR-based method incorporates bi-directional contexts and exploits the global expressivity of Transformer to capture long-range relationships on sequential tokenized information. 

Our main contributions are summarized as follows,
\begin{itemize}
    \item We propose a novel framework M6-Fashion to conduct multi-modal image generation and editing.
    \item M6-Fashion cooperates both the high-fidelity generation ability from StyleGAN-based method and high-flexibility sequence modeling ability in one unified framework.
    \item M6-Fashion generalizes and achieve the state-of-the-art performance on a various image generation and editing tasks. It is of great potential to server as an efficient AI designer for real Fashion industry.
\end{itemize}

%% file: Tables/method-comparision.tex
\begin{table*}[t]
\caption{Method comparisons between representative image synthesis models.}
\vspace{-4mm}
\label{table:method-comparision}
\begin{center}
{
\begin{tabular}{|l|l||c|c|c|c|}
\rowcolor{LightGray}
\hline
Paradigm & Method & Explicit Prior & Spatial-aware & Multi-modal & Non-Autoregressive \\ \hline\hline
\multirow{1}{*}{VAE-based} & VDVAE~\cite{vdvae} & \xmark & \cmark & \xmark & \cmark  \\ \hline\hline
\multirow{1}{*}{Diffusion-based} & DDPM~\cite{ddpm} & \xmark & \cmark & \xmark & \cmark \\ \hline\hline
\multirow{1}{*}{GAN-based} & StyleGAN~\cite{stylegan2} & \cmark & \xmark & \xmark & \cmark \\ \hline\hline
\multirow{2}{*}{Transformer-based} & Taming Transformer~\cite{esser2021taming} & \xmark & \cmark & \cmark & \xmark \\ \cline{2-6}
& \textbf{M6-Fashion (Ours)} & \cmark & \cmark & \cmark & \cmark \\
\hline
\end{tabular}
}
\end{center}
\end{table*}

%% file: Sections/2-related-work.tex
\noindent\textbf{GAN-based Conditional Image Synthesis.}
BigGAN \cite{biggan} trains a generator conditioned on semantic label. StyleGAN \cite{stylegan} introduces a style-based generator to produce high-resolution photo-realistic images in a specific domain, such as face, cat and others. TediGAN \cite{xia2020tedigan}  use different encoders to convert multi-modal information into a common latent space of StyleGAN. To conduct different tasks, it adopt different extra models, which is inconvenient and difficult to handle cross-modal control signals. StyleCLIP \cite{styleclip} applies a pretrained visual-linguistic similarity model to optimize the latent code for different text descriptions, which is time-consuming. ContraGAN \cite{contragan} adopts contrastive learning to supervise the relevance between images and text descriptions. The  aforementioned GAN-based Image synthesis methods rely on different extra models to conduct different tasks.

\noindent\textbf{VQ-based Conditional Image synthesis.}
Recently, VQ-based two-stage image synthesis methods stand out in visual applications~\cite{li2020consistent, li2020meta, li2020self, li2021super}. DALLE \cite{dalle} builds a transformer framework that autoregressively models the text and image tokens as a single stream of data. The discrete variational autoencoder is trained to compress each 256×256 RGB image into a 32×32 grid of image tokens. VQGAN \cite{esser2021taming} introduces an adversarial loss to improve the perceptual quality of the codebook. Cogview \cite{ding2021cogview} investigates the potential of finetuning diverse downstream tasks. M6-UFC \cite{zhang2021m6} explores the advantage of  unifying the multi-modal control signals. Aforementioned methods train the discrete variational autoencoder to reconstruct the images in global domain, which ignores the style prior knowledge in a specific domain, such as face or fashion clothes. Besides, these works try to increase the codebook size to keep good perceptual quality, which leads to highly cost in memory. In contrast, our framework trains spatial-aware Style-based generator and use a product quantizer to learn style prior knowledge and generate an exponentially large codebook at low cost in time and memory.


\noindent\textbf{Non-Autoregressive Sequence Modeling.} 
Though it is natural to autoregressively predict tokens from left to right when generating a sequence, autoregressive decoding suffers from the slow speed and
sequential error accumulation issues. 
Thus, the non-autoregressive generation (NAR)~\cite{ghazvininejad2019mask,guo2020incorporating,liao2020probabilistically,mansimov2019generalized} paradigm is proposed to avoid these drawbacks in neural machine translation, image captioning, and speech synthesis. 
These approaches often employ the bidirectional Transformer (i.e. BERT) as it is not trained with a specific generation order. 
Our progressive NAR generation algorithm improves upon the Mask-Predict non-autoregressive algorithm~\cite{zhang2021m6}, by introducing the to facilitate sample selection and dynamic termination.

%% file: Sections/3-method.tex
\begin{figure*}[t]
\centering
\includegraphics[width=\textwidth]{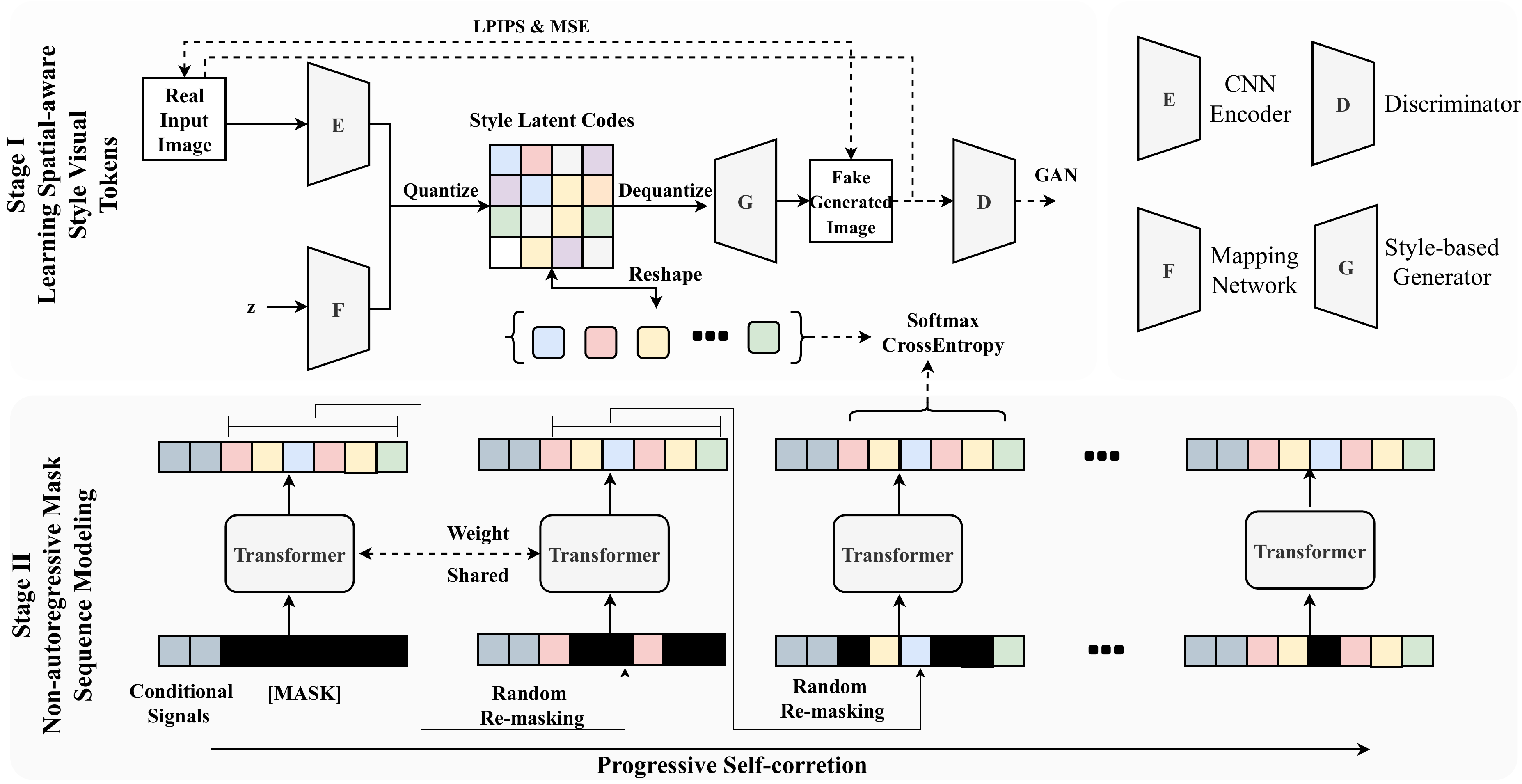}
\vspace{-2mm}
\caption{An overview of our proposed two-stage transformer-based M6-Fashion framework. In stage I (\S\ref{sec:learn-discrete-representation}), we quantize an image into spatial-aware style visual tokens. In stage II (\S\ref{sec:multi-modal-image-sythesis}, given conditional information, we predict these visual tokens in a non-autoregressive sequence modeling manner.
M6-Fashion inherits both the high-fidelity generation ability from StyleGAN-based method and high-flexibility sequence modeling ability in one unified framework.  
}
\label{fig:framework}
\end{figure*}

\noindent\textbf{Overview.}
Our goal is to explore the domain-specific prior knowledge and unify the multi-modal controls for image generation and editing.
As shown in Figure~\ref{fig:framework}, we propose a simple yet effective Transformer-based two-stage framework, namely, M6-Fashion.
In the first stage, we learn the discrete latent representations for converting an image into a sequence of discrete tokens with explicit spatial dimensions. (\S\ref{sec:learn-discrete-representation})
Our proposed discrete representation benefits from its spatial-aware characteristic, enabling the generative adversarial networks (GANs) to effectively encode the local semantics of an image into its latent representations.
In the second stage, we adopts a non-autoregressive (NAR) Transformer-based network to capture the distribution over sequences of discrete tokens, comprising of tokens from multi-modal control signals and style visual tokens from the target image. (\S\ref{sec:multi-modal-image-sythesis}) 
Our NAR-based method incorporates bidirectional contexts and exploits the global expressivity of Transformer to capture long-range relationships on sequential tokenized data.
Benefiting from our effective framework design, M6-Fashion can be applied to various conditional image synthesis tasks as shown in Figure~\ref{fig:teaser}. (\S\ref{sec:unify-application})  
Our key insight to obtain an effective and expressive framework is that, it takes the domain-specific style knowledge of GANs and retains the flexibility of transformer together, to achieve high-fidelity AI-fashion image synthesis tasks.

\subsection{Learning spatial-aware discrete latent codes}\label{sec:learn-discrete-representation}

\noindent\textbf{Discovering Spatial-aware Style Prior Knowledge.}
Style-based generator architectures~\cite{stylegan,stylegan2} have achieved supreme performance on image synthesis tasks.
The key to the success of these style-based generators lies in exploring the \emph{style prior knowledge} based on the intermediate latent space. 
These style-based generators choose an intermediate latent style representations $\mathbf{z}$, which then controls the style of the image at each layer via Adaptive Instance Normalization (AdaIN).
It has been demonstrated that such a design allows a less entangled representation learning in $\mathcal{z}$, leading to better generative image modeling.
Despite the promising results, the style-based generator can suffer from the global style latent codes, losing fine-grained local control for the images.
This global style representation also limits the applications (\textit{e.g.} local editing, local mixing), which requires spatially manipulating the image.
In contrast, instead of learning a vector-based latent representation $\mathbf{z} \in \mathcal{R}^{d}$, we utilize a tensor with explicit spatial dimensions $\mathbf{z} \in \mathcal{R}^{h \times w}$.
Our proposed style latent code benefits from its decoupled spatial dimensions, enabling the model to easily encode the local semantics of images into the representation space. 
By discovering the spatial-aware prior knowledge, our method also offers a novel capability to edit specific regions of an image by manipulating the corresponding positions in the latent codes $\mathbf{z}$.


\noindent\textbf{Generate Discrete Style Visual Tokens.}
To utilize the highly expressive transformer architecture for both conditional and unconditional image synthesis tasks, we first need to express the aforementioned style representation $\mathbf{z}$ in the form of a tokenized sequence, \textit{i.e.} style visual tokens.
In order to generate the discrete style visual tokens, a codebook $\mathcal{Z}=\left\{\mathbf{z}_{k}\right\}_{k=1}^{K}$ for style vector quantization is learned, where $\mathbf{z}_{k} \in \mathbb{R}^{n_{z}}$ is the $k$-th code-word in the codebook and $K$ is the number of code-words.
Any image $\mathbf{x} \in \mathbb{R}^{H \times W \times 3}$ can be represented by a spatial collection of codebook entries $\mathbf{Z} \in \mathbb{R}^{h \times w \times n_{z}}$.
Concretely, a convolutional encoder $E$ first encodes the original image $\mathbf{x}$ as $\hat{\mathbf{z}}=E(\mathbf{x}) \in \mathbb{R}^{h \times w \times n_{z}}$.
Then an element-wise quantization operation $\mathbf{q}(\cdot)$ is applied to each element $\hat{z}_{i j}$ to obtain the element’s closest code-word $z_{k}$, following

\begin{equation}\label{eq:quantization}
z_{\mathbf{q}}=\mathbf{q}(\hat{z}):=\left(\underset{z_{k} \in \mathcal{Z}}{\arg \min }\left\|\hat{z}_{i j}-z_{k}\right\|\right) \in \mathbb{R}^{h \times w \times n_{z}}.
\end{equation}

To reconstruct the input image $\mathbf{x}$, a style-based generator $G$ is learned for recovering image $\mathbf{x} \in \mathbb{R}^{H \times W \times 3}$ from $z_{\mathbf{q}}$ such that $\hat{\mathbf{x}}$ is close to $\mathbf{x}$.

\begin{equation}
\hat{\mathbf{x}}=G\left(z_{\mathbf{q}}\right)=G(\mathbf{q}(E(x)))
\end{equation}

To effectively learn such a codebook $\mathcal{Z}$ to generate style visual tokens, we adopt a straight-through gradient estimator~\cite{bengio2013estimating} to achieve back propagation through the non-differentiable quantization operation $\mathbf{q}(\cdot)$. Concisely, we copy the gradients from the decoder $G$ to the encoder $E$, such that the model and codebook can be trained end-to-end via the loss function,

\begin{equation}
\begin{aligned}
\mathcal{L}_{\mathrm{quant}}(E, G, \mathcal{Z})=\|x-\hat{x}\|^{2} &+\left\|\operatorname{sg}[E(x)]-z_{\mathbf{q}}\right\|_{2}^{2} \\
&+\left\|\operatorname{sg}\left[z_{\mathbf{q}}\right]-E(x)\right\|_{2}^{2},
\end{aligned}
\end{equation}

where $\operatorname{sg}[\cdot]$ denotes the stop-gradient operation.

In this way, we turn the style latent codes $\hat{\mathbf{z}}$ into a sequence of discrete codes $I \in \{0, \ldots,|\mathcal{Z}|-1\}^{h \times w}$, where each index specifies the respective entries in the learned codebook.

\noindent\textbf{Building Context-rich Vocabularies via Product Quantization.}
Using transformers to represent images as discrete distribution over latent style tokens requires us to learn rich codebook and push the limits of compression rate. 
To do so, former work~\cite{dalle,ding2021cogview,esser2021taming} use a simple straight-forward quantization operation $\mathbf{q}(\cdot)$, and try to keep good perceptual quality by increasing the codebook size $|\mathcal{Z}|$. 
However, to achieve better compression rate, we adopt an effective vector quantization method inspired by Product quantization (PQ)~\cite{jegou2010product}.
A product quantizer can generate an exponentially large codebook at very low time and memory cost. 
We decompose the high-dimensional style latent code $\hat{\mathbf{z}}$ into the Cartesian product of subspaces and then quantize these subspaces, separately.
Formally, denote the embedding $\hat{z} \in \mathcal{R}^{n_z}$ as the concatenation of $M$ sub-vectors: $\hat{z}=\left[\hat{z}^{1}, \ldots \hat{z}^{m}, \ldots \hat{z}^{M}\right]$. 
For simplicity all the sub-vectors have common number of dimensions $n_z/M$.
The whole quantization process can be split into $M$ separate sub-problems, each of which can be solved following
Eq.\eqref{eq:quantization}.
The final quantized latent vector is the concatenation $z_{\mathbf{q}}=\left[z_{\mathbf{q}}^{1}, \ldots z_{\mathbf{q}}^{m}, \ldots z_{\mathbf{q}}^{M}\right]$.


The benefit of product quantization is that it can efficiently generate a codebook $\mathcal{Z}$ with a large number of code words. 
If each sub-codebook has $K$ sub-code words, then their Cartesian product has in equivalent to $K^M$ code words .
In this way, we achieve lower quantization distortions, more computationally efficient, and better image synthesis quality.
We qualitatively compare the effect of different quantization alternatives in Figure~\ref{fig:ablation-quantization}.

\subsection{Multi-modal Image Generation and Editing}\label{sec:multi-modal-image-sythesis}
 As depicted in Figure~\ref{fig:framework}, our M6-Fashion modifies the original BERT~\cite{devlin2018bert} model to accommodate various multi-modal controls using non-autoregressive sequence modeling. 

\noindent\textbf{Multi-modal Conditional Signals.} 
In various synthesis tasks, the user may provide additional information $\mathbf{c}$ to control the generation process, \textit{i.e.}, how an example shall be synthesized.
This conditional information $\mathbf{c}$, could be a single label class, a text describing the overall image, or even another reference image itself.
To fully embrace the flexibility of transformer architecture,
we conduct the conditional image synthesis by tokenizing a set of control signals $\mathbf{c}$.
In our work, we summarize the conditional control signals into three major aspects.
As described in \S\ref{sec:learn-discrete-representation}, we can covert an image into a sequence of code-words (i.e. style visual tokens), based on stage one’s encoder $E$ and codebook $\mathcal{Z}$.
Hence, the Visual Condition (VC) is denoted by a sequence $\mathbf{v} \in \{0, \ldots, |\mathcal{Z}| - 1 \}^{N_v}$ consisting of code words from the codebook $\mathcal{Z}$, where $N_{v}$ is the sequence length. 
A Textual Condition (TC) consists of a sequence of words $\mathbf{t} \in \{0, \ldots, |\mathcal{W}| - 1 \}^{N_t}$, where $\mathcal{W}$ is the textual vocabulary and $N_t$ is the number of words in the text. 
A Preservation Condition (PC), it is a sequence of binary masks $\mathbf{p} \in \{0, 1\}^{N_p}$ with the same length as the visual tokens for the image, where 1 means that the token is known while 0 means the token needs to be predicted. 
In our M6-Fashion, we aim to synthesize the target image’s sequence $I$ conditioned on multiple multi-modal condition $\mathbf{c}$, \textit{i.e.}, a combination of control signals from $\{\mathbf{v}, \mathbf{t}, \mathbf{p}\}$.

\begin{figure}[t]
\centering
\includegraphics[width=\linewidth]{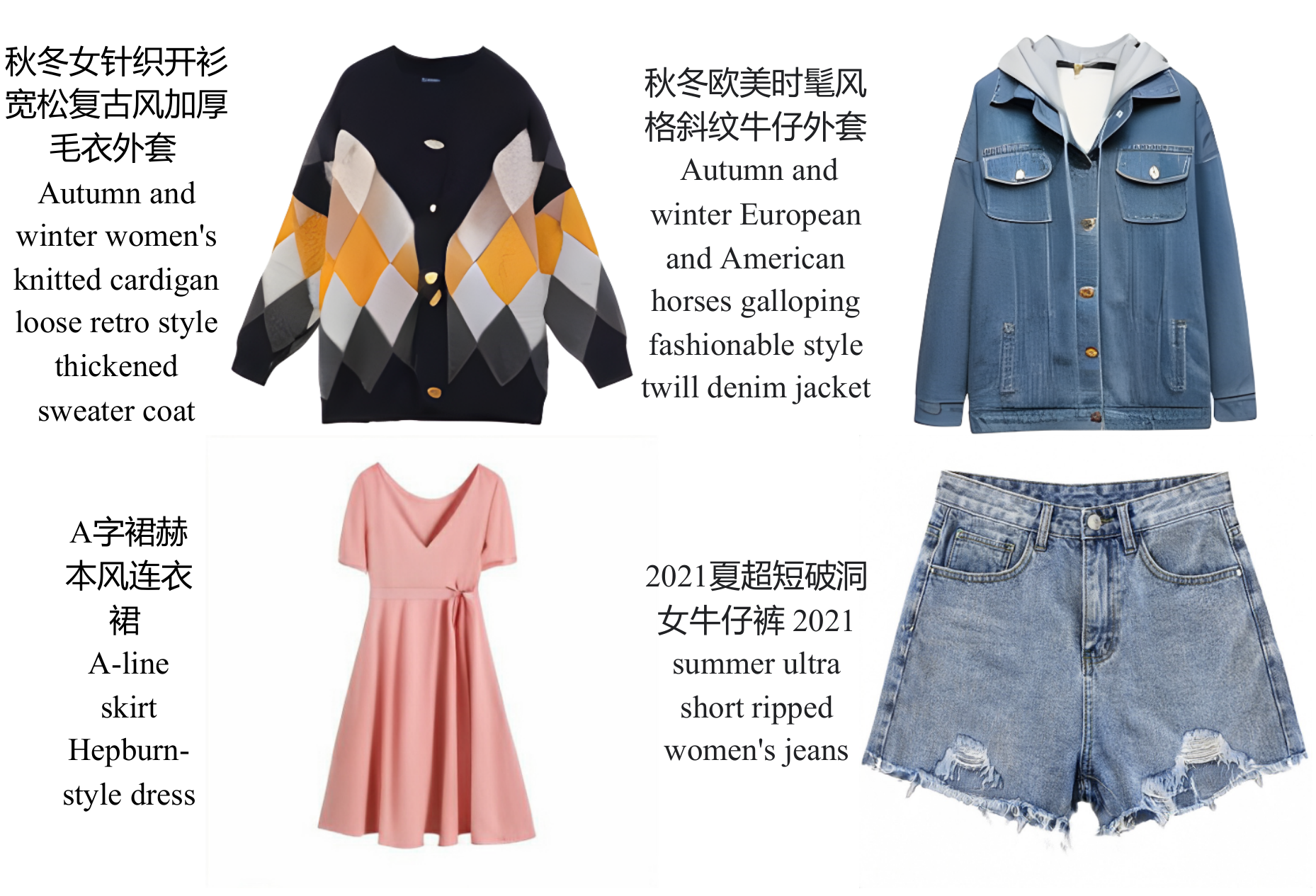}
\vspace{-2mm}
\caption{High-resolution 1024x1024 generated images. Best viewed zoomed in.}
\label{fig:hr-image}
\end{figure}

\noindent\textbf{Non-Autoregressive Sequence Modeling.}
Existing two-stage works~\cite{dalle,esser2021taming} adopts the AutoRegressive (AR) paradigm to conduct sequence modeling. 
However, the AR paradigm suffers from slow generation speed, cannot fully support preservation control signals and fails to capture bidirectional contexts. 
To tackle the shortcoming of AR, in our M6-Fashion design, we adopt a novel Non-AutoRegressive (NAR) approach to unify various number of multi-modal controls.

Our M6-Fashion adopts the original BERT model~\cite{devlin2018bert} to accommodate the multi-modal image generation and editing tasks.
The backbone model is a multi-layer bidirectional Transformer encoder, enabling the dependency modeling between all input elements. 
The input sequence of M6-Fashion starts with multi-modal condition tokens $\mathbf{c}$ and ends with the visual token sequence $\mathbf{I}$ of the target image to generate.
Two special separation tokens [EOV] and [EOT] are appended to the end of the visual condition and text condition sequences, respectively. 
If there are multiple visual controls, another special token [SEP] is inserted to separate them.
The sequence $\mathbf{I}$ of the target image to generate may be partially or fully masked by a special token [MASK]. 
When the preservation condition is present and $\mathbf{p}_i = 1$, token $\mathbf{I}_i$ is set to the code-word corresponding to the given image block to be preserved.
We train our M6-Fashion via masked sequence modeling, \text{i.e.}, predicting the masked tokens in the target image conditioned on the controls $\mathbf{c}$.
The task is defined as learning the likelihood of the sequence given the conditional information $\mathbf{c}$,

\begin{equation}
p(\mathbf{I} | \mathbf{c})=\prod_{i \in M } p\left(I_{i} | \mathbf{I}_{U}, \mathbf{c} \right),
\end{equation}

where $M$ is the set of masked positions in the target image while $U$ is the unmasked counterparts.

We optimize the NAR sequence modeling by minimizing the softmax cross-entropy loss defined as,

\begin{equation}
\mathcal{L}_{\text {NAR}}=\mathbb{E}_{i \sim M}[-\log p\left(I_{i} | \mathbf{I}_{U}, \mathbf{c} \right)].
\end{equation}

\begin{figure}[t]
\centering
\includegraphics[width=\linewidth]{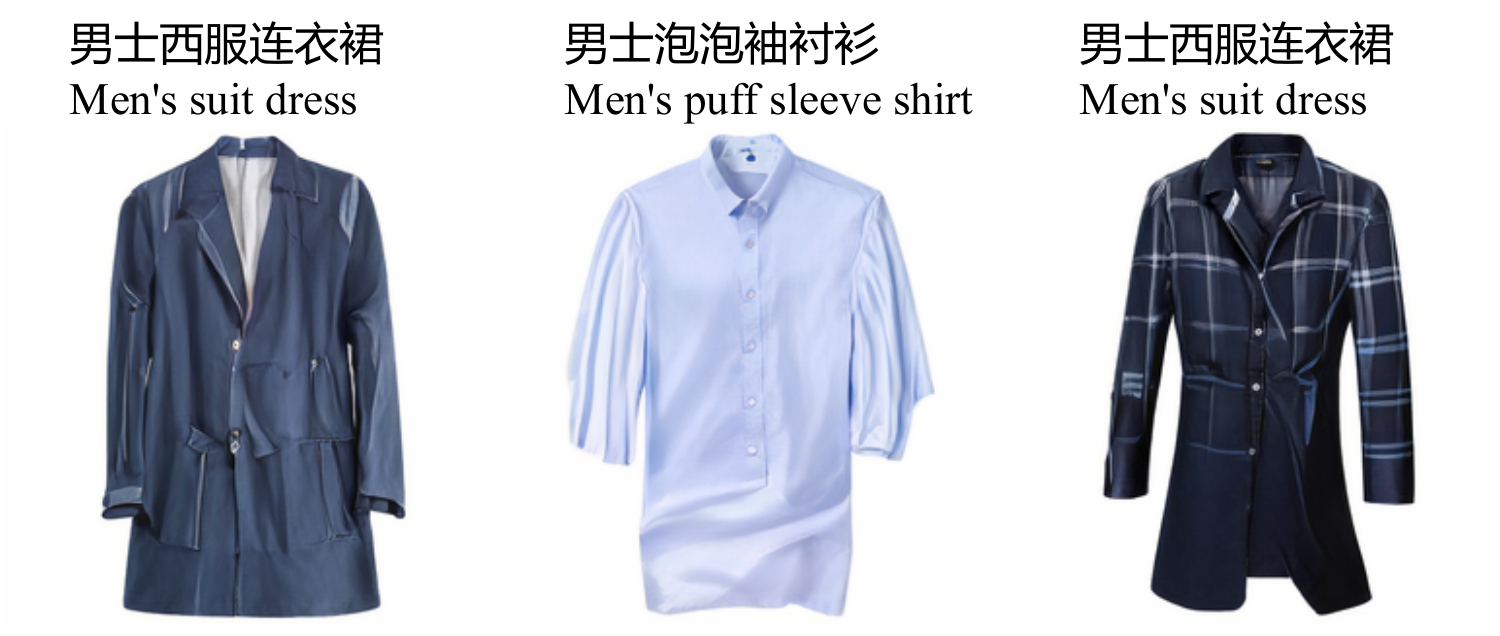}
\vspace{-2mm}
\caption{Our M6-Fashion shows strong zero-shot generation ability to synthesize counterfactual samples that do not exist in real world.}
\label{fig:counterfactual-samples}
\end{figure}

\noindent\textbf{Self-correction for Masked non-AutoRegressive Training.}
We train M6-Fashion via masked sequence modeling, \textit{i.e.}, predicting the masked visual tokens of the target image conditioned on the conditional signals.
This task is similar as the Masked Language Modeling~\cite{devlin2018bert}, but incorporates multi-modal conditional signals when predicting the masked tokens.

Given a fully-masked sequence at the first iteration, we predict all the target tokens.
Formally, the visual tokens in $\mathbf{I}^{(0, \text{out})}$ for $t = 0$ is all set to [MASK], except for the positions that are controlled by the preservation signals $\mathbf{p}$.
Then, we iteratively re-mask and re-predict a subset of tokens with a constant number of iterations. 

Each iteration of our SMART algorithm consists of a \textit{mask step} and then a \textit{predict step}. 
During the mask step at iteration $t$, we first remain $N_I -n $ visual tokens unchanged, which are sampled from a Multinomial distribution proportional to the probability scores $p\left(I_{i} | \mathbf{I}_{U}, \mathbf{c} \right)$.
And all the other $n$ tokens are re-masked to [MASK]. 
We gradually decreases the number of tokens to re-mask for every iteration, following $n=N_{I} \cdot\left(\beta+\frac{T-t}{T-1} \cdot(\alpha-\beta)\right)$, where $\alpha$ is the initial mask ratio, $\beta$ is the minimum mask ratio, and $T$ is the maximum possible number of iterations.
During the prediction step, Given the condition $\mathbf{c}$ and an former re-masked sequence $\mathbf{I}^{(t, \text{in})}$, M6-Fashion predicts a distribution $p\left(I_{i} | \mathbf{I}_{U}, \mathbf{c} \right)$ for each masked position $i$.
We illustrate the effect of progressive mask sequence modeling in Figure~\ref{fig:smart-per-step}.

\begin{figure*}[t]
\centering
\includegraphics[width=\textwidth]{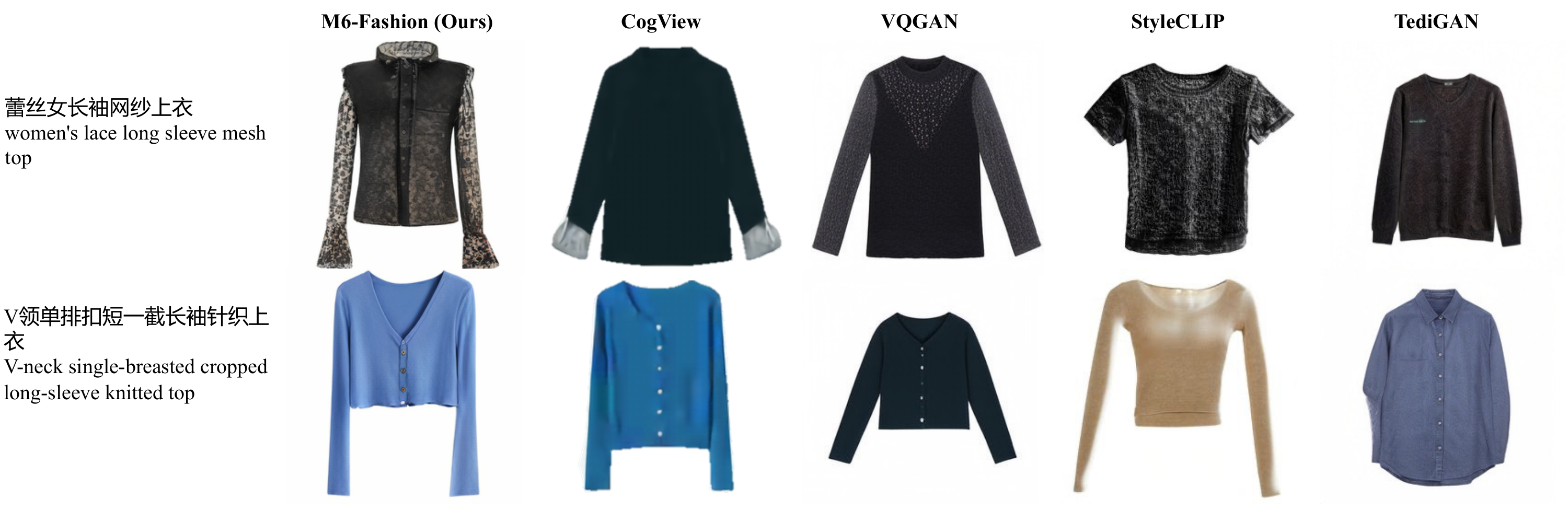}
\label{fig:t2i-compare}
\vspace{-2mm}
\caption{Qualitative comparisons with state-of-the-art methods on text-to-image generation task.}
\end{figure*}

\subsection{Unified framework for various AI-Fashion tasks}\label{sec:unify-application}
As shown in Figure~\ref{fig:teaser}, the goal of our M6-Fashion framework is to construct a unified framework for various AI-Fashion tasks, including but not limited to, i) the unconditional image generation task, ii) the text-to-image generation task, iii) the local mixing task, iv) the text-guided local editing task, v) the image in-painting/out-painting task and vi) the style-mixing task.

To accommodate a wide variety of potential downstream applications, we construct training samples by mask parts of the target image $\mathbf{I}$ to predict using four sub-strategies: (1) randomly mask the desired number of tokens, (2) mask all tokens, (3) mask the tokens within some boxed areas of the image, where the number of boxes and the box sizes are randomly sampled, (4) mask the tokens outside some random boxed areas of the image. 
Experimentally, we use the four strategies with probability 0.70, 0.10, 0.10, and 0.10, respectively.

To accommodate a wide variety of potential multi-modal conditional signal $\mathbf{c}$ for each training sample, we choose four different combinations: $\{\mathbf{v},\mathbf{t}\}$, $\{\mathbf{v}\}$, $\{\mathbf{t}\}$, $\{\varnothing\}$, where $\{\mathbf{v},\mathbf{t}\}$ means the visual and textual conditions are simultaneously employed, $\{\mathbf{v}\}$ or $\{\mathbf{t}\}$ means only a visual or textual condition signal is used, and $\{\varnothing\}$ means the unconditional situation.
Note that the preservation condition is already included in the masked sequence modeling task. 
Since our dataset does not contain ground-truth pairs of visual controls and target images, we crop one or multiple regions of a target image to construct $\{\mathbf{v}\}$ for the target image $\mathbf{I}$.
Because the text-to-image synthesis from solely textual controls is more challenging, we force the textual condition situation get more attention.
Experimentally, we choose the four combinations with probability 0.20, 0.55, 0.20, 0.05, respectively.


%% file: Sections/4-experiment.tex
\input{Tables/text2img-sota-comparision}

\subsection{Experimental Settings}
\noindent\textbf{Datasets.} 
In this paper, we mainly focus on the practical field of AI-Fashion, which has a wide application for e-commerce and fashion design.
Accordingly, we collect a very large-scale clothing dataset M2C-Fashion~\cite{zhang2021m6} with text descriptions, which contains tens of millions of image-text pairs, much larger than the commonly used general text-to-image datasets COCO~\cite{lin2014microsoft}.

\noindent\textbf{Evaluation metrics.} 
We consider two commonly used evaluation metrics for image synthesis tasks. 
Inception Score (IS)~\cite{salimans2016improved} measures both the confidence of the integral of the marginal probability of the predicted classes (diversity) and the conditional class predictions for each synthetic image (quality).
Higher scores mean that the model can generate more distinct images.
Fréchet Inception Distance (FID)~\cite{heusel2017gans} compares the statistics (mean and variances of Gaussian distributions) between the generated samples and real samples.
Lower scores indicate that the model can generate higher quality images.
For the text conditional generation tasks, we additionally calculate the CLIP-Score~\cite{radford2021learning} to measure the similarity between the text descriptions and the generated images.

\noindent\textbf{Implementation details.} 
For the BERT model~\cite{devlin2018bert} in our M6-Fashion, we set the number of layers, hidden size, and the number of attention heads to 24, 1024, and 16, respectively. 
Our M6-Fashion has $307M$ parameters.
The model capacity is same as the transformer used by other baseline models. 
In the first stage, similar as~\cite{kim2021exploiting}, we choose a CNN-based model as the encoder model and a StyleGAN-based as the decoder model.
Following the experimental settings~\cite{esser2021taming}, we use the 256x256 image size and quantize each image into a discrete sequence of 16x16 codes.
We set the group of product quantization as 4, where in each group, the codebook size $|\mathcal{Z}_i|$ is set to 256.
Thus, the total codebook storage cost remains the same as other baseline models.
As for hyper-parameters of SMART, we set the initial mask ratio $\alpha$, the minimum mask ratio $\beta$, and the maximum iteration number $T$ to 0.8, 0.2, and 10, respectively.
We may initialize M6-Fashion with a large-scale pre-trained network, \textit{e.g.}, M6~\cite{lin2021m6}. However, to ensure fair comparisons with other baseline methods, we train M6-Fashion from scratch using only the training dataset.

\begin{figure*}[t]
\centering
\includegraphics[width=\linewidth]{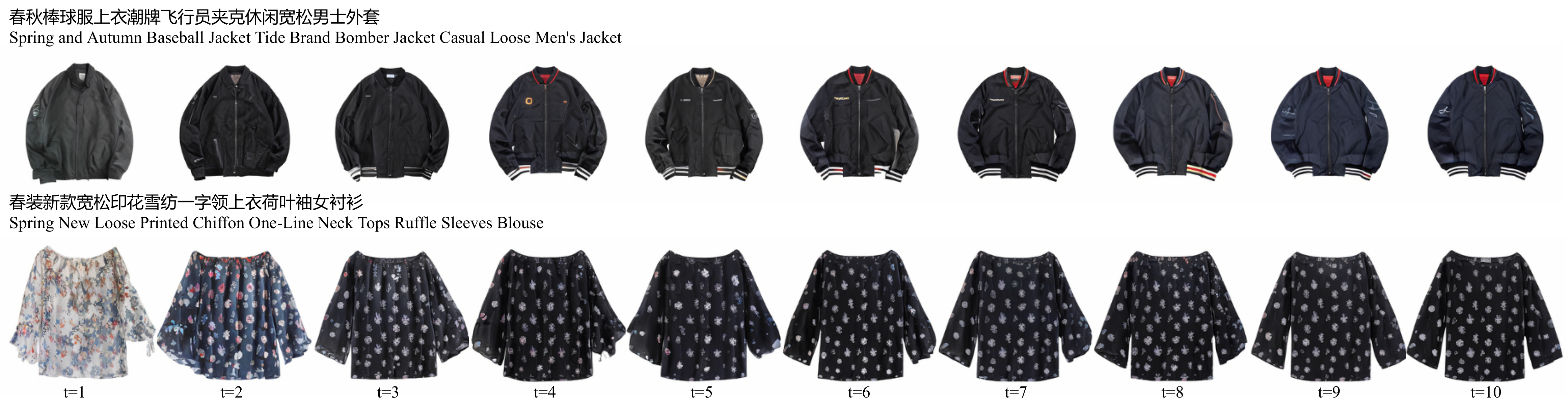}
\vspace{-2mm}
\caption{The iterative inference process of our SMART algorithm.}
\label{fig:smart-per-step}
\end{figure*}

\subsection{A Unified Framework for Image Synthesis Tasks}
In our M6-Fashion design, we adopt the transformer architecture due to its generality, flexibility and versatility, making it a promising architecture for multi-modal conditional control.
In Figure~\ref{fig:teaser}, we qualitatively verify the synthesis ability of M6-Fashion and perform experiments on various conditional image task, including:
\begin{enumerate}
\item \textbf{Unconditional image generation task}. We randomly sample and generate the images without any unconditional information. Results are depicted in Figure~\ref{fig:unconditinoal-samples}. 
\item \textbf{Text-to-image generation task}, where the we condition on the text description of M2C-Fashion. The resulting text-to-image samples are visualized in Figure~\ref{fig:text2img-samples}.
\item \textbf{Local editing task}. Given a local pattern image as reference, we locally edit the corresponding area of the original image. Figure~\ref{fig:local-editing-samples} shows that M6-Fashion generate meaningful designs by migrating the local pattern from reference image to original image.
\item \textbf{Text-guided local editing task}. We generate design images given both text description and local pattern reference as conditional information. 
\item \textbf{In-painting/Out-painting task}, where we complete a mixing area in original image. Results are demonstrated in Figure~\ref{fig:in-out-painting-samples}.
\item \textbf{Style-mixing task}. We gradually mix the style between two reference images and generate intermediate results as shown in Figure~\ref{fig:styl-mixing-samples}.
\end{enumerate}
Instead of requiring a specific architecture or model for each task, M6-Fashion allows us to unify any combinations of multi-modal controls to synthesize high-fidelity images. 
The proposed approach M6-Fashion is an unified, efficient, general framework for conditional image generation and editing. 
Furthermore, M6-Fashion supports one or multiple visual controls for more flexible synthesis, as shown in Figure~\ref{fig:teaser} where we generate images given multiple visual conditions.
We observe that M6-Fashion can reasonably aggregate multiple visual elements and produce a harmonious fashion clothing image.
In our experiments, M6-Fashion outputs images with a resolution of 256x256 pixels to ensure fair comparisons with existing methods.
However, benefiting from the fast inference speed of NAR, M6-Fashion enables image synthesis beyond a resolution of 256×256 pixels.
Impressive high-resolution results (1024x1024 pixels) are demonstrated in Figure~\ref{fig:hr-image}.


\input{Tables/uncondition-sota-comparision}

\subsection{Comparison with SOTA methods for Text-to-Image Synthesis}
In this section, we investigate how our M6-Fashion quantitatively compares to other existing state-of-the-art models. 
We select the most common and challenging task, \textit{i.e.}, text-to-image synthesis to judge the image generation ability.

First, We compare our M6-Fashion with state-of-the-art  methods in Table~\ref{table:text2image-sota-comparision}, including StyleGAN-based text-to-image models TediGAN~\cite{xia2020tedigan} and StyleCLIP~\cite{styleclip} and Transformer-based CogView~\cite{ding2021cogview} and VQGAN~\cite{esser2021taming}.
Our M6-Fashion achieves the best performance on both FID and IS score, even outperforming the competitor TediGAN that uses complex instance-level optimization.
Also as shown in Figure.5, the visualization results well demonstrates design of M6-Fashion are suitable for text-to-image conditional synthesis.
M6-Fashion aggregates both the high-fidelity image generation ability from GAN-based model and the flexibility sequence modeling ability from Transformer-based model,

Besides, we compare the inference speed of M6-Fashion with other methods in Table~\ref{table:text2image-sota-comparision}.
Our M6-Fashion achieves X8 times speedup compared to other two-stage transformer-based method VQGAN and CogView.
This suggests that our non-autoregressive M6-Fashion with SMART generation algorithm can achieve high-fidelity images with a fast inference speed.

Further, in Figure~\ref{fig:counterfactual-samples}, we intuitively investigate the zero-shot generation ability, \textit{i.e.}, the ability to generate counterfactual cases. 
We find that M6-Fashion can synthesize high-quality images, even for the counterfactual case, while other methods may potentially fail.

\subsection{Comparison with SOTA methods for Unconditional Image Generation}

In this section, we quantitatively investigate how our M6-Fashion model compares to existing models for unconditional generative image synthesis. 
Particularly, in Table~\ref{table:uncondition-sota-comparision}, we assess the performance of our model in terms of FID and compare to a variety of representative models for different generative paradigms (VAEs~\cite{vdvae}, GANs~\cite{stylegan2,kim2021stylemapgan}, Flows~\cite{ddpm}, AR~\cite{esser2021taming}).
Although some highly-customized GAN model reports a slightly better FID scores, our approach provides a unified model that works well across a variety of tasks while retaining the ability to accommodate any conditional information. 

\subsection{Ablation study}
In our ablation study, unless otherwise stated, we conduct all the experiments on text-to-image synthesis task on M2C-Fashion dataset.

\begin{figure}[t]
\centering
\includegraphics[width=\linewidth]{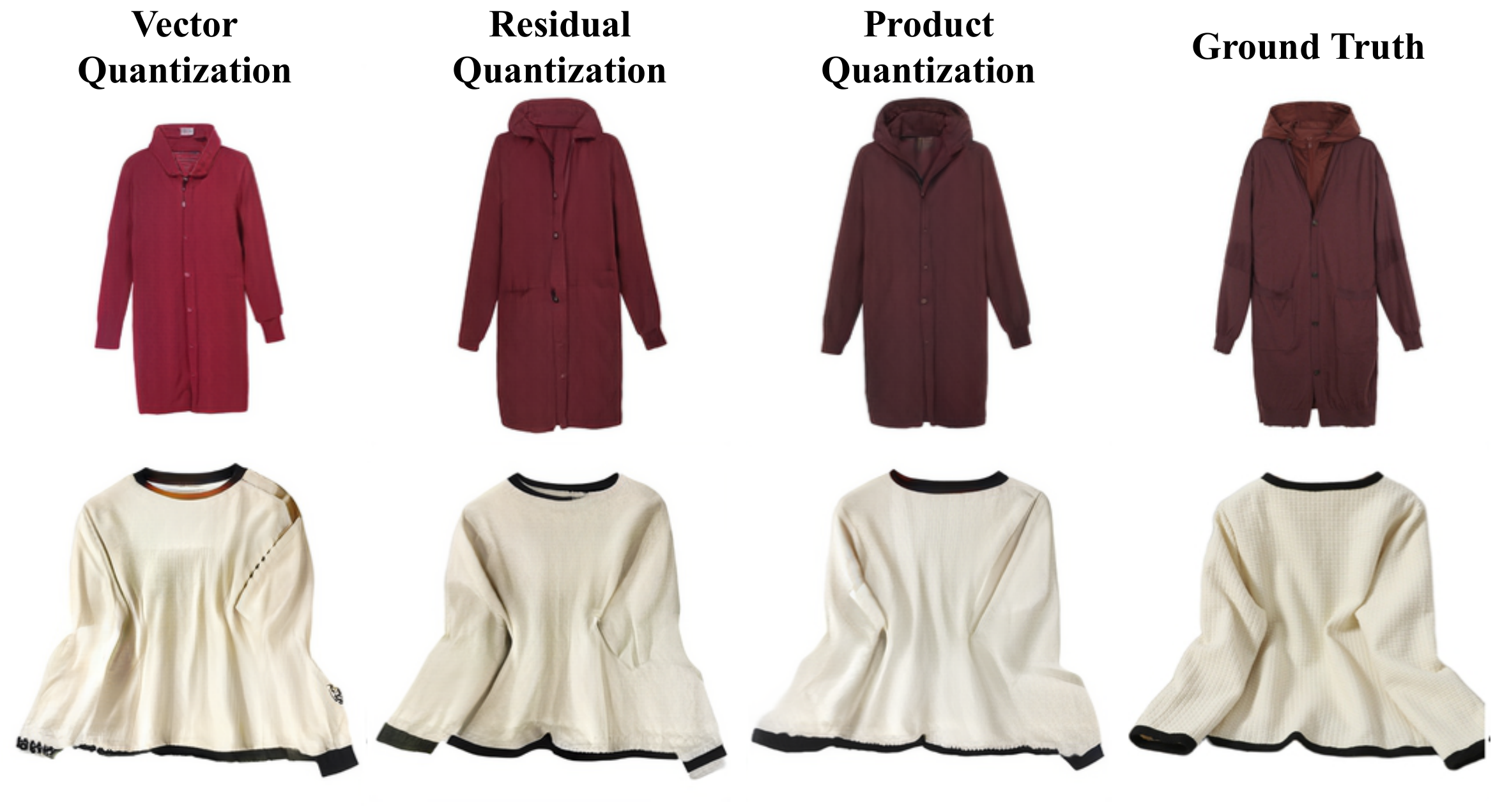}
\vspace{-2mm}
\caption{Our product quantization achieves high-quality, style-consistent image reconstruction results.}
\label{fig:ablation-quantization}
\end{figure}

\noindent\textbf{Low-distortion discrete representation learning.}
In our first stage, we quantize an image into a discrete representation as style visual tokens.
However, the quantization operation may inevitably introduce information losses.
In M6-Fashion design, we aim to achieve low-distortion discrete representation as much as possible.
Here, we analyze the effectiveness of our PQ (\S\ref{sec:learn-discrete-representation}) compared with other quantization alternatives, including simple vector quantization used in, residual quantization.
As shown in Table~\ref{table:ablation-quantization}, our PQ strategy achieves better synthesis scores within the same codebook storage size.
The visualizations in Figure~\ref{fig:ablation-quantization} shows that our PQ achieves more style-consistent reconstruction images.
All these results demonstrate the effectiveness of the low-distortion production quantization operation.

\input{Tables/ablation-quantization}

\noindent\textbf{Progressive NAR Generation Algorithm.}
In Figure~\ref{fig:smart-per-step}, we visualize in the iterative process of our SMART inference method.
We can find that the text-image relevance and fidelity of the generated images gradually increase as the iteration progresses, verifying our SMART algorithm can guide the inference process towards a better direction and synthesize more realistic images given the conditional information.

\subsection{Applications beyond AI-Fashion}
\begin{figure}[t]
\centering
\includegraphics[width=\linewidth]{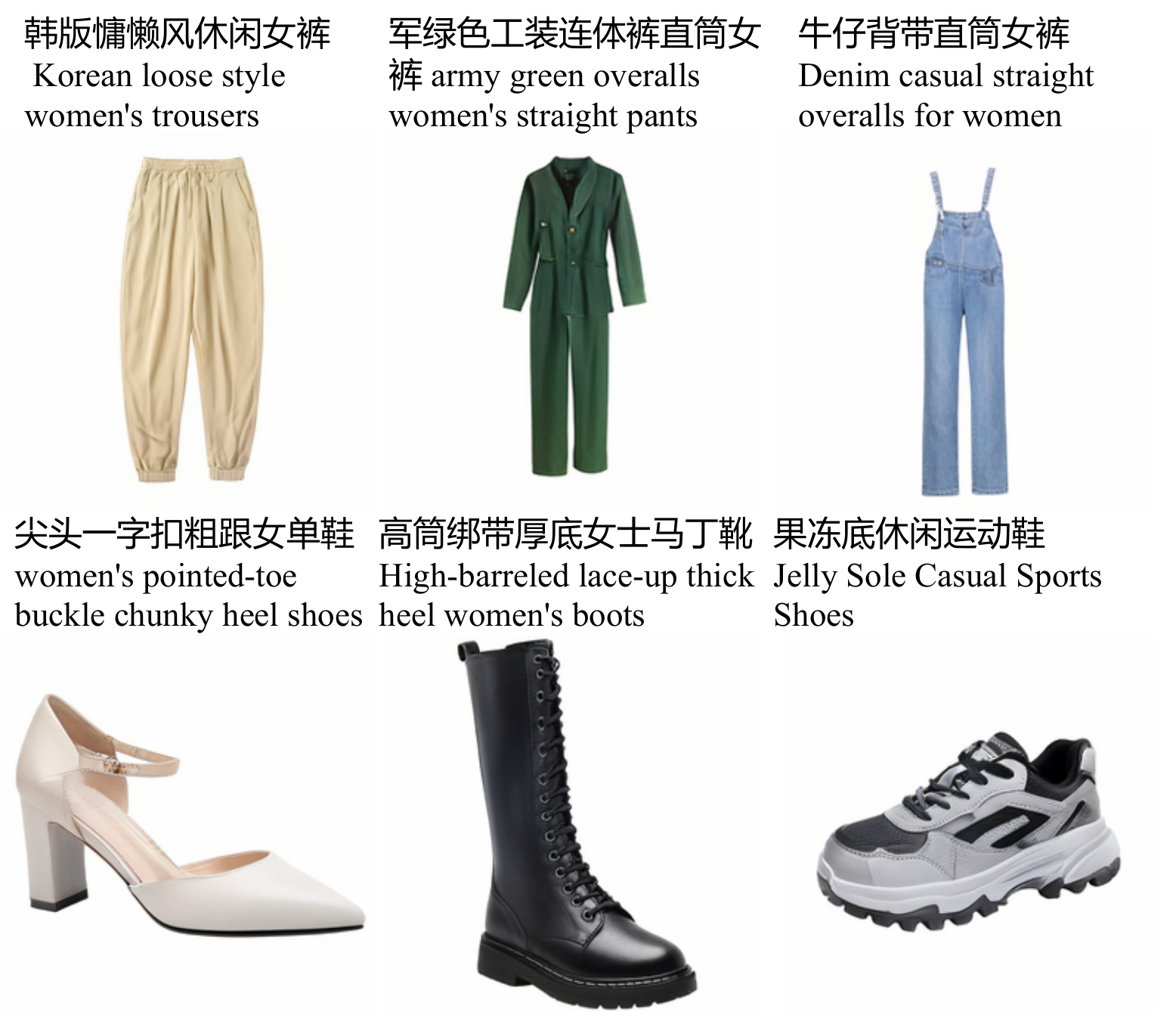}
\vspace{-2mm}
\caption{A wide application potential for M6-Fashion.}
\label{fig:other-application}
\end{figure}

Although our M6-Fashion is initially designed and focused on AI-fashion field, however, the proposed method can be seamlessly adapted to other domain-specific fields, \textit{e.g.,} footwear appearance design, cosmetic appearance design.
We show some qualitative results for these applications in Figure~\ref{fig:other-application}.
This shows the flexible and versatile application potential of the M6-Fashion framework.

%% file: Tables/text2img-sota-comparision.tex
\begin{table}[t]
\label{tabel:text-to-image-comparision}
\caption{Comparisons for the text-to-image conditional generation task on the M2C-Fashion \emph{test} set. $\uparrow$ indicates the higher the better, and $\downarrow$ the lower the better.}
\vspace{-4mm}
\label{table:text2image-sota-comparision}
\begin{center}
{
\begin{tabular}{|l||c|c|c|c|}
\rowcolor{LightGray}
\hline
Method & FID $\downarrow$ & IS $\uparrow$ & CLIP-score $\uparrow$ & Speed  \\ \hline\hline
TediGAN~\cite{xia2020tedigan} & $39.93$ & $2.85$ & $0.30$ & 98s  \\ \cline{1-5} 
StyleCLIP~\cite{styleclip} & $31.17$ & $4.01$ & $0.31$ & 20min \\
\hline\hline
CogView~\cite{ding2021cogview} & $39.79$ & $3.42$ & $0.32$ & 34.8s  \\ \cline{1-5} 
Taming Transformer~\cite{esser2021taming} & $10.28$ & $3.82$ & $0.38$ & 8.7s \\ \cline{1-5} 
\textbf{M6-Fashion (Ours)} & $\mathbf{5.69}$ & $\mathbf{4.49}$ & $\mathbf{0.38}$ & 1.02s \\ \hline
\end{tabular}
}
\end{center}
\end{table}

%% file: Tables/uncondition-sota-comparision.tex
\begin{table}[t]
\caption{Comparisons for the unconditional generation task. $\uparrow$ indicates the higher the better, and $\downarrow$ the lower the better.}
\vspace{-4mm}
\label{table:uncondition-sota-comparision}
\begin{center}
{
\begin{tabular}{|l|l||c|c|}
\rowcolor{LightGray}
\hline
& Method & FID $\downarrow$ & IS $\uparrow$ \\ \hline\hline
\multirow{1}{*}{Diffusion-based} & DDPM~\cite{ddpm} & $47.63$ & $2.71$ \\ \hline\hline
\multirow{2}{*}{GAN-based} & StyleGAN~\cite{stylegan2} & $10.06$ & $4.35$ \\ \cline{2-4} 
& StyleMapGAN~\cite{kim2021stylemapgan} & $16.43$ & $3.87$ \\
\hline\hline
\multirow{2}{*}{Transformer-based} & Taming Transformer~\cite{esser2021taming} & $30.76$ & $3.17$ \\ \cline{2-4}
& \textbf{M6-Fashion (Ours)} & $\mathbf{12.98}$ & $\mathbf{4.14}$ \\
\hline
\end{tabular}
}
\end{center}
\end{table}

%% file: Tables/ablation-quantization.tex
\begin{table}[h]
\caption{Ablations for different quantization methods.}
\vspace{-4mm}
\label{table:ablation-quantization}
\begin{center}
{
\begin{tabular}{|l||c|c|}
\rowcolor{LightGray}
\hline
Method & FID $\downarrow$ & IS $\uparrow$ \\ \hline\hline
Vector Quantization & $6.59$ & $4.36$ \\
Residual Quantization & $5.45$ & $4.21$ \\ \hline\hline
Product Quantization & $\mathbf{4.38}$ & $\mathbf{4.16}$ \\
\hline
\end{tabular}
}
\end{center}
\end{table}

%% file: Sections/conclusion.tex
We proposed M6-Fashion to unify free-form combinations of multi-modal conditional controls in a universal form for image synthesis. 
We utilized non-autoregressive generation to improve inference speed, enhance holistic consistency, and support preservation controls. 
Further, we designed a progressive generation algorithm based on relevance and fidelity estimators to ensure relevance and fidelity. 
In the future, we will explore the universal synthesis capabilities of M6-Fashion on open domains based on large-scale pre-trained models like M6.

%% file: Sections/appendix.tex
\begin{figure*}[t]
\centering
\includegraphics[width=\textwidth]{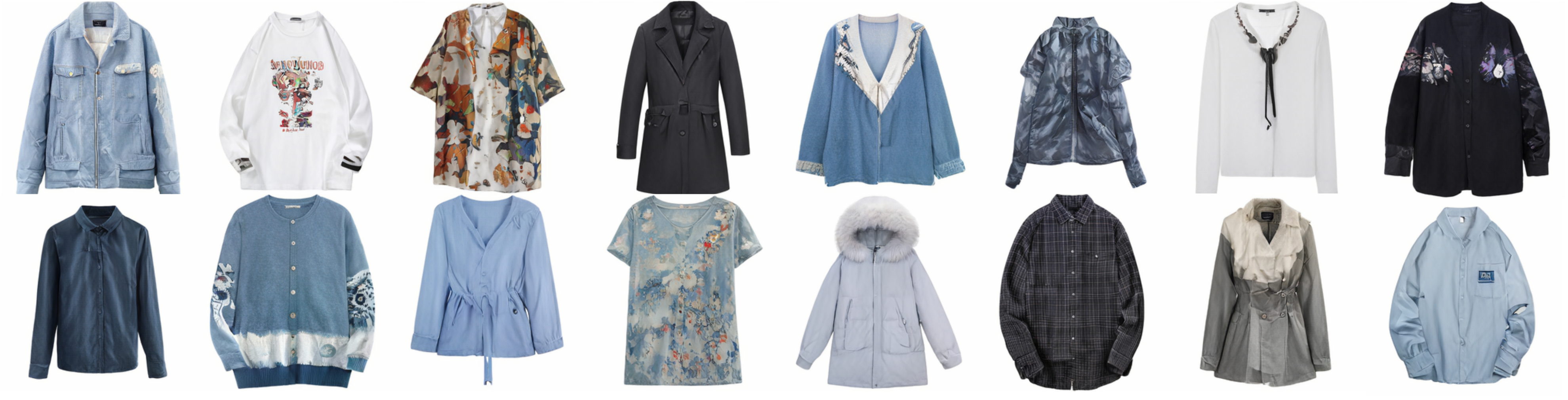}
\caption{More qualitative results for the unconditional image generation task.}
\label{fig:unconditinoal-samples}
\end{figure*}
\begin{figure*}[t]
\centering
\includegraphics[width=\textwidth]{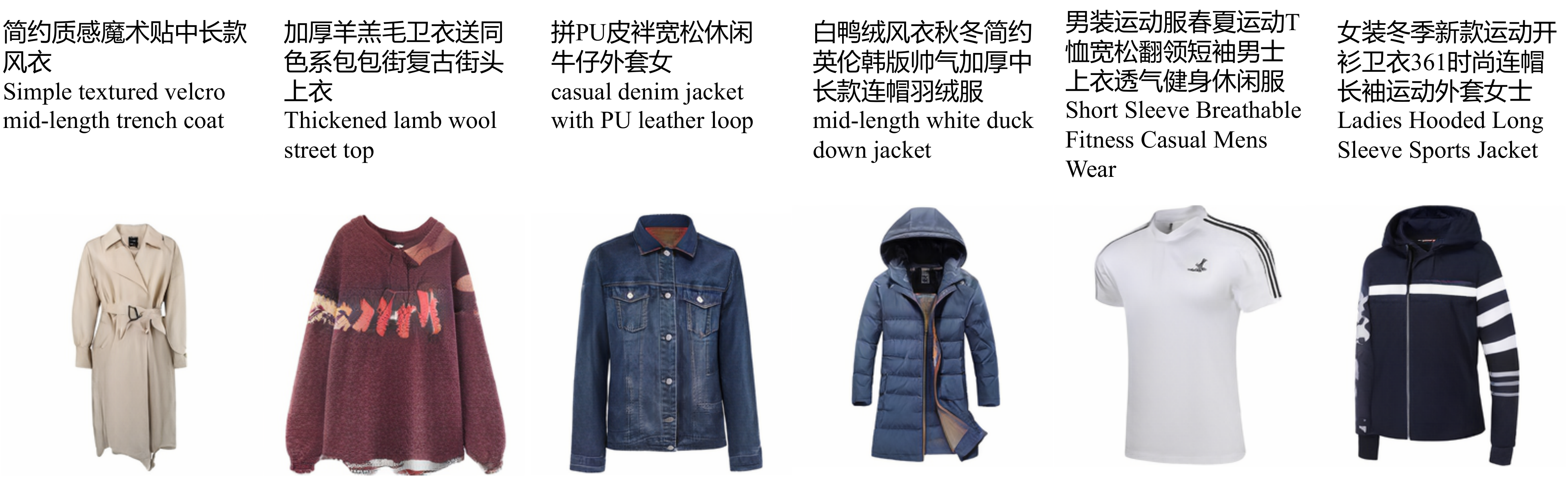}
\caption{More qualitative results for the text-to-image generation task.}
\label{fig:text2img-samples}
\end{figure*}
\begin{figure*}[t]
\centering
\includegraphics[width=0.95\textwidth]{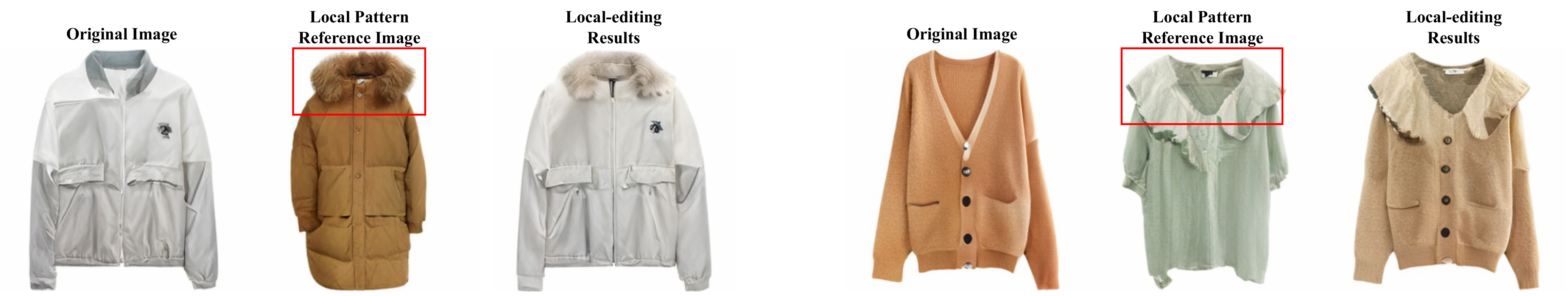}
\caption{More qualitative results for local editing task.}
\label{fig:local-editing-samples}
\end{figure*}
\begin{figure*}[t]
\centering
\includegraphics[width=\textwidth]{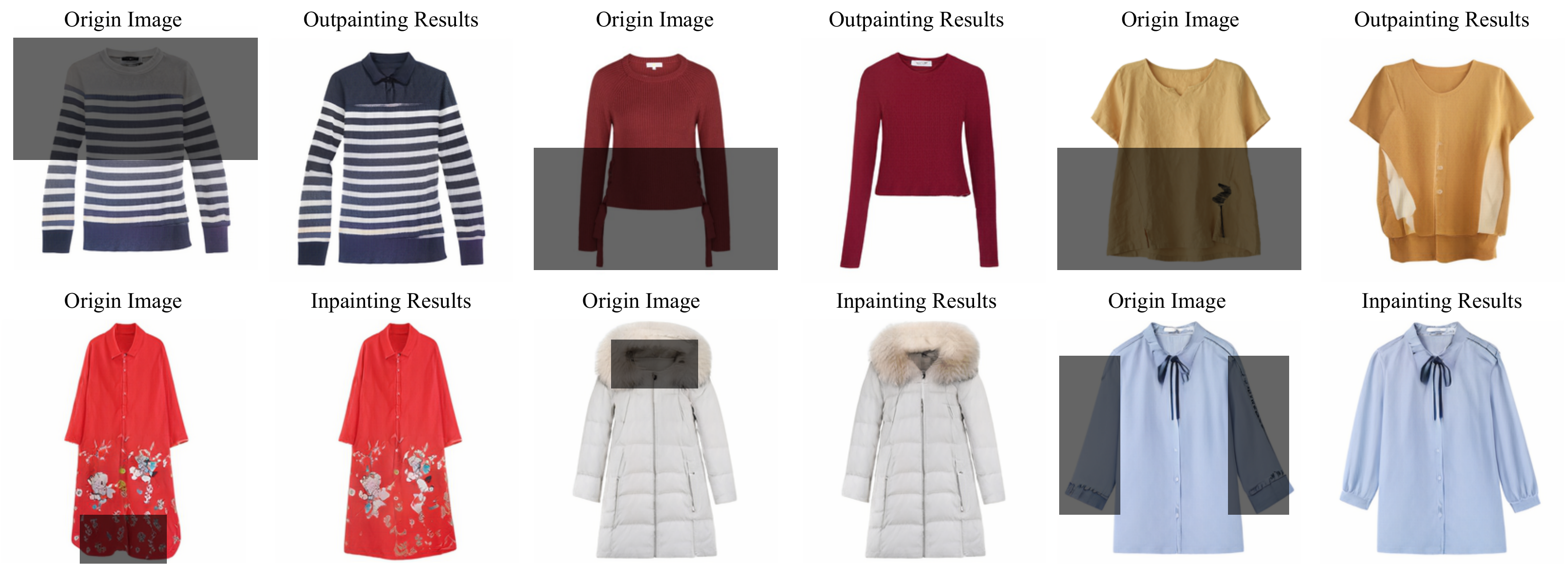}
\caption{More qualitative results for in-painting and out-painting task. Masked area are illustrated in  half-transparent black boxes.}
\label{fig:in-out-painting-samples}
\end{figure*}
\begin{figure*}[t]
\centering
\includegraphics[width=0.8\textwidth]{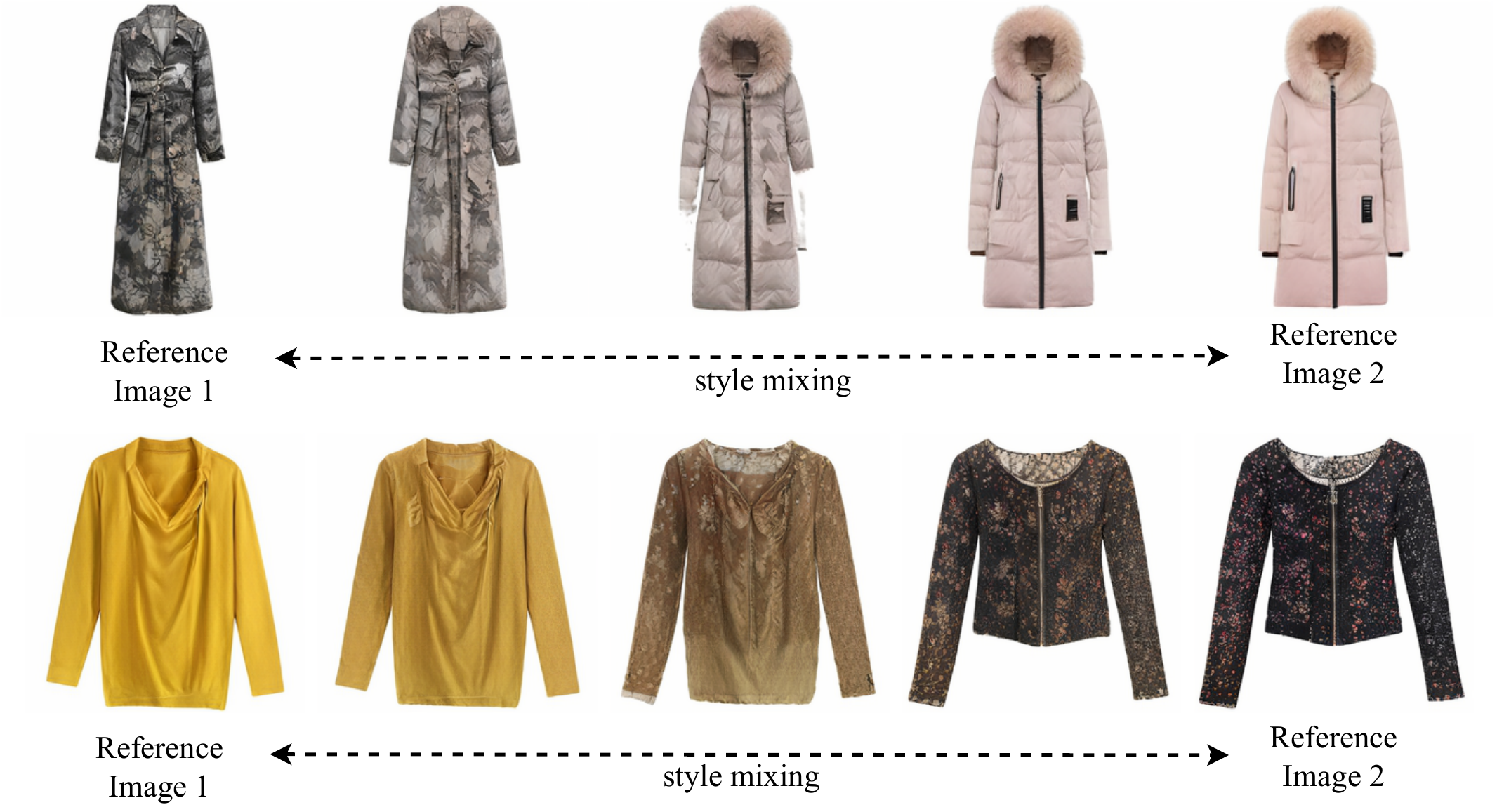}
\caption{More qualitative results for the style mixing task.}
\label{fig:styl-mixing-samples}
\end{figure*}